\pdfoutput=1

\documentclass[11pt]{article}

\usepackage[final]{acl}

\usepackage{times}
\usepackage{latexsym}

\usepackage[T1]{fontenc}

\usepackage[utf8]{inputenc}

\newcommand{\dataset}{\textsc{CiteCheck}\xspace}

\usepackage{microtype}

\usepackage{inconsolata}

\usepackage{graphicx}

\usepackage{amsthm,amsmath,amssymb}
\usepackage{bbding}
\usepackage{float}
\usepackage{bm}
\usepackage{booktabs}
\usepackage{graphicx}
\usepackage{multirow}
\usepackage{diagbox}
\usepackage{enumitem}
\usepackage{setspace}
\usepackage{algorithm}
\usepackage{algorithmic}
\usepackage{soul}
\usepackage{CJKutf8}
\usepackage{xspace}

\title{\dataset: Towards Accurate Citation Faithfulness Detection}

\author{Ziyao Xu$^1$, Shaohang Wei$^1$, Zhuoheng Han$^1$, Jing Jin$^1$, Zhe Yang$^1$\\ \textbf{Xiaoguang Li$^2$, Haochen Tan$^2$, Zhijiang Guo$^2$, Houfeng Wang$^1$}\\
$^1$National Key Laboratory for Multimedia Information Processing, \\
School of Computer Science, Peking University \\
$^2$Huawei Noah’s Ark Lab \\
\texttt{\{xzyxzy, 11jj617, yz\_young, wanghf\}@pku.edu.cn} \\
\texttt{\{shaohang, 2100017789\}@stu.pku.edu.cn \quad lixiaoguang11@huawei.com} \\
\texttt{haochetan2-c@my.cityu.edu.hk \quad zhijiangguo@hkust-gz.edu.cn}
}

\begin{document}
\maketitle
\begin{abstract}

Citation faithfulness detection is critical for enhancing retrieval-augmented generation (RAG) systems, yet large-scale Chinese datasets for this task are scarce. Existing methods face prohibitive costs due to the need for manually annotated negative samples. To address this, we introduce the first large-scale Chinese dataset \textsc{CiteCheck} for citation faithfulness detection, constructed via a cost-effective approach using two-stage manual annotation. This method balances positive and negative samples while significantly reducing annotation expenses. \textsc{CiteCheck} comprises training and test splits. Experiments demonstrate that: (1) the test samples are highly challenging, with even state-of-the-art LLMs failing to achieve high accuracy; and (2) training data augmented with LLM-generated negative samples enables smaller models to attain strong performance using parameter-efficient fine-tuning. \textsc{CiteCheck} provides a robust foundation for advancing citation faithfulness detection in Chinese RAG systems. The dataset is publicly available at \url{https://github.com/xzy-xzy/CiteCheck} to facilitate research.
\end{abstract}

\section{Introduction}
Large Language Models (LLMs) are prone to generating factual errors through hallucinations when answering real-world questions. Retrieval-augmented generation (RAG) systems \cite{DBLP:conf/nips/LewisPPPKGKLYR020,GuuLTPC20,BorgeaudMHCRM0L22} address this limitation by leveraging external information retrieval to ground LLM responses in verifiable sources. Recent advancements extend RAG systems to generate text with inline citations \cite{DBLP:conf/emnlp/GaoYYC23}, enabling users to validate the reliability of generated content by cross-referencing cited documents. However, studies reveal a critical weakness in these systems: citation faithfulness. A substantial portion of generated text may lack proper support from the cited references \cite{DBLP:conf/emnlp/LiuZL23, hu2024evaluate}, undermining the trustworthiness and verification capability of RAG outputs. This challenge necessitates accurate citation faithfulness detection—determining whether cited passages genuinely support their associated claims—as a fundamental requirement for improving RAG reliability.

Developing robust citation faithfulness detection methods requires large-scale, high-quality datasets. While English benchmarks have emerged \cite{DBLP:conf/emnlp/YueWCZS023}, Chinese datasets remain notably absent. Constructing such resources presents unique challenges: realistic negative samples (unsupported citations) must originate from strong RAG systems to ensure practical usage, yet these systems rarely produce such errors. For instance, a RAG system with a 10\% error rate would require annotating approximately 70,000 samples to collect 7,000 negative examples—a prohibitively expensive endeavor. This tension between dataset quality and construction cost demands innovative solutions for efficient data curation without compromising sample integrity.

To bridge this gap, we introduce \textsc{CiteCheck}, the first large-scale Chinese dataset for citation faithfulness detection. Our approach combines 11,307 knowledge-intensive questions with a novel two-stage annotation framework that reduces labeling costs while preserving data quality. \textsc{CiteCheck} comprises two distinct components designed to address both detection difficulty and training efficacy. 

The development and test sets each contain 500 positive (supported) and 500 negative (unsupported) samples totaling 2,000 unmodified RAG outputs. Experimental analysis demonstrates these original samples pose significant challenges, with state-of-the-art LLMs achieving limited detection accuracy. The training set includes 9,796 samples (4,898 positive/negative pairs) where negative instances are generated through LLM-based document modification rather than relying solely on rare RAG errors. Despite this augmentation, parameter-efficient fine-tuning on 7B-8B parameter models yields strong detection performance, confirming the preserved quality of modified negative samples.

Our contributions are threefold:  \textsc{CiteCheck} establishes the first comprehensive benchmark for Chinese citation faithfulness detection; (2) We propose an efficient data augmentation strategy that reduces annotation costs by 86\% compared to conventional approaches; (3) Extensive experiments validate the dataset’s quality and utility, showing that models trained on our augmented data effectively generalize to challenging real-world samples. This work advances reliable RAG development by providing essential resources and methodologies for building verifiable, citation-grounded LLM applications in Chinese.

\section{Dataset Construction}
\begin{table*}
\scriptsize
\centering

\begin{tabular}{|p{0.07\textwidth}|p{0.2\textwidth}|p{0.2\textwidth}|p{0.2\textwidth}|p{0.2\textwidth}|}
\hline
Question & 
\begin{CJK}{UTF8}{gbsn} 
特斯拉在中国的纯电动汽车销量占比是多少？
\end{CJK} 
\par What is Tesla's share of all-electric car sales in China?
& 
\begin{CJK}{UTF8}{gbsn} 
乘坐飞机的时候托运一个行李箱，再带一个20寸的箱子，带上飞机的箱子会被称重吗？
\end{CJK} 
\par When I check a suitcase on an airplane and bring a 20” case, will the case be weighed when I bring it on the plane?
& 
\begin{CJK}{UTF8}{gbsn} 
为什么当你把一幅画切成几块时，每一块都只有一部分，但是当你把一张全息图切成几块时，每一块都有完整的图像？
\end{CJK} 
\par Why is it that when you cut a painting into pieces, each piece has only a part of it, but when you cut a hologram into pieces, each piece has the full image?
& 
\begin{CJK}{UTF8}{gbsn} 
如果闪电劈中混凝土大楼之类的，就算在房间里处于隔缘状态，也会受伤害么？
\end{CJK} 
\par If lightning strikes a concrete building or something like that, can you get hurt even if you're in a room that's insulated?

\\ \hline

\par Answer \par (\underline{Statement}) & 
\begin{CJK}{UTF8}{gbsn} 
\underline{\textcolor{red}{特斯拉在2023年上半年纯电动}}
\underline{\textcolor{red}{汽车市场的占有率为21.7\%。}}
\end{CJK}
\par \textcolor{red}{Tesla's share of the all-electric car market in the first half of 2023 is 21.7 percent.}
& 
\begin{CJK}{UTF8}{gbsn} 
是的，乘坐飞机时...。
\underline{\textcolor{red}{每件托}}
\underline{\textcolor{red}{运行李的重量不能超过50公斤}}
\underline{\textcolor{red}{，而每位旅客随身携带物品的}}
\underline{\textcolor{red}{重量以5公斤为限。}}
此外，每位旅客...。
\end{CJK}
\par \textcolor{red}{The weight of each piece of checked baggage cannot exceed 50 kilograms, while the weight of each passenger's carry-on is limited to 5 kilograms.}
& 
\begin{CJK}{UTF8}{gbsn} 
这是因为...。
\underline{全息的原理在于}
\underline{将动量空间通过全息技术记录}
\underline{在实空间里，因此，\textcolor{red}{如果全息}}
\underline{\textcolor{red}{图缺失一部分，那部分实空间}}
\underline{\textcolor{red}{的信息就丢失了，像就不完整}}
\underline{\textcolor{red}{了。}}
即使...。
\end{CJK}
\par \textcolor{red}{If a portion of the hologram is missing, that portion of real space is lost and the likeness is incomplete.}
&

\begin{CJK}{UTF8}{gbsn} 
闪电击中混凝土大楼...
\underline{在澳}
\underline{大利亚新南威尔士州纽卡斯尔}
\underline{的一个案例中，闪电击中了一}
\underline{座大楼，虽然\textcolor{red}{大楼的结构保持}}
\underline{\textcolor{red}{良好}，但是巨大的爆裂声和震}
\underline{动可能会对内部的人造成伤害}
\underline{。}因此，...。

\end{CJK}

\par \textcolor{red}{The building is structurally sound.}

\\ \hline

\par Cited \par Documents & 
\begin{CJK}{UTF8}{gbsn}
[1]【\textcolor{blue}{2023上半年}全球纯电动汽车销量出炉...】...据该报道，\textcolor{blue}{特斯拉在纯电动汽车市场期间占据21.7\%的份额。}...
\end{CJK} 
\par \textcolor{blue}{First half of 2023 / Tesla held a 21.7\% share of the all-electric car market during the period.}
&
\begin{CJK}{UTF8}{gbsn}
\par [1] 办理\textcolor{blue}{托运行李}对行李物品规定如下：...\textcolor{blue}{每件行李物品重量不能超过50公斤。}...
\par \textcolor{blue}{Check-in baggage / The weight of each baggage item can not exceed 50 kilograms.}
\par [2] \textcolor{blue}{随身携带物品的重量，每位旅客以5公斤为限。}...
\par \textcolor{blue}{The weight of carry-on items is limited to 5 kg per passenger.}
\end{CJK} 
&
\begin{CJK}{UTF8}{gbsn}
\par [1] ...\textcolor{blue}{如果普通照片缺失一部分，那部分实空间的信息就丢失了，像就不完整了。}全息照片如果缺失一部分，同样会造成信息的缺失，但是...
\end{CJK} 
\par \textcolor{blue}{If a portion of an ordinary photograph is missing, that portion of real space is lost and the likeness is incomplete.}
&
\begin{CJK}{UTF8}{gbsn}
\par [1] ...澳大利亚新南威尔士州纽卡斯尔...可清楚看到闪电击中大楼的场面，同时可听到巨大的爆裂声。据悉，闪电所击中的大楼为一处健身房。...
\end{CJK} 

\\ \hline

Label & positive & positive & negative & negative \\
\hline

Note & supported by a single document & supported by multiple documents & contradictory information & unmentioned information \\

\hline
\end{tabular}
\caption{\label{tab:dataset} Sample examples of the dataset. For the answer and cited documents we show only part of the content. We underline the selected statement in the answer. We mark in \textcolor{red}{red} and \textcolor{blue}{blue} the key information associated with the label in the statement and the cited documents. We provide English translations of the questions and key information.}
\vspace{-0.4cm}
\end{table*}

\subsection{Question Collection}
We collect Chinese questions from the sources:
\noindent \textbf{WebText} \cite{bright_xu_2019_3402023}: A large-scale Chinese community question-answering dataset spanning diverse topics.

\noindent \textbf{WebCPM} \cite{DBLP:conf/acl/QinCJYLZLHDWXQL23}: A Chinese long-form question-answering dataset focused on interactive web search contexts.

\noindent \textbf{Zhihu-KOL} \cite{zhihu-kol}: A high-quality question-answering dataset derived from Zhihu, a prominent Chinese QA platform.

\noindent \textbf{RGB} \cite{DBLP:conf/aaai/0011LH024}: A bilingual question-answering dataset based on news reports.

\noindent \textbf{TrickQA}: Questions with ambiguous, incorrect, or unverifiable premises (see Appendix~\ref{app:question} for details).

After collecting these questions, we input them into an open-sourced RAG system to simulate real-world question-answering scenarios and analyze how the system processes and responds to these diverse inputs. The RAG system retrieves five external documents and generates responses. Statements in the answers are annotated with citation marks (1–5), indicating alignment with information from the corresponding documents. On average, each statement spans 33.4 tokens, while each document averages 177.3 tokens. An original sample is formed by pairing a labeled statement with its cited documents, represented as a tuple (question, answer, statement, cited documents).

\subsection{Data Augmentation}

The goal of data augmentation is to create negative samples of high quality by making minor modifications to the cited documents in the original samples. Given the use of an industrial RAG system, the number of negative samples in the original samples is estimated to be insufficient. To construct a balanced training set, as well as a label-balanced dev set and test set for evaluation, successfully augmented negative samples can be used. The modified documents should not be inconsistent or incoherent, so as not to provide the trained model with a false basis for judging the negative samples.

We use GPT-4o \cite{openai2024gpt4technicalreport} for data augmentation. After providing the original sample to the LLM, it is asked to perform the following steps in sequence:

\noindent \textbf{Segments Identification}: Find all key segments in the cited document that directly support the information in the statement.

\noindent \textbf{Segments Grouping}: Group the key segments by the information they support, with each group containing key segments that support the same or related information in the statement.

\noindent \textbf{Segments Modification}: Select a group of key segments and modify them so that they do not support the corresponding information in the statement. 

The modification changes only the portion that relates to the supported information in the statement. This maintains logical flow and non-contradictory information within the key segments, and keep the key segments logical in the context of the document and non-contradictory to other information in the document. If there is more than one key segment in a group, the information in all of them should be consistent after the modification.

For each sample, the LLM is asked to try two methods of modification:

\noindent\textbf{Content Revision}: Alter specific details within a key segment without introducing direct contradictions to the original information.

\noindent \textbf{Structure Preservation}: Remove information from a key segment while ensuring the overall coherence and integrity of the segment remain intact.

After completing the LLM augmentation, each original sample is accompanied by the LLM-labeled key segment information and corresponds to the two augmented samples generated by the LLM using the two modification methods. The cost is 0.026\$ per sample. See Appendix~\ref{app:aug} for more details of the augmentation.

\subsection{Two-stage Manual Annotation}

The original samples need to be manually labeled as positive or negative samples before they can be used to form the dataset (examples are shown in Table~\ref{tab:dataset}). In the LLM augmentation phase, although we try to guide the LLM to augment negative samples with qualified quality, the LLM may generate some samples that do not meet the requirements. Therefore, the augmented samples also need to be manually labeled for compliance before they can be used to form the dataset. The goal of the two-stage manual annotation is to complete the manual annotation needed above.

In the first stage, the annotators (from the professional data annotation institution in China) need to label whether the original sample is a positive or negative sample, i.e., to determine whether the sum of the information provided by the cited documents fully supports the statement. In order to reduce the difficulty of labeling, the information of key segments labeled by LLM will be provided to the annotators as a reference. However, since the LLM labeling is not always accurate, if the annotators are unable to make a judgment after reading the key segments, they still need to read other parts of the documents to make a judgment. 
In this stage, the number of negative samples identified by the annotation is 1,006, with a negative sample rate of about 9\%. We randomly selected 2,000 samples (1,000 negative and 1,000 positive) and split them equally to create the development and test sets. The augmented samples corresponding to the positive samples in the remaining original samples will be labeled in the second stage.

In the second stage, the annotators need to determine whether an augmented sample is of acceptable quality and whether it is a negative sample. In order to reduce the difficulty of labeling, we show the annotator a comparison of the documents before and after the modification in the form of modification traces. Among the augmented samples that the annotators determine to be negative samples of acceptable quality, we select 2,449 samples that use the modification methods of changing information and deleting information respectively, totaling 4,898 samples. These augmented negative samples together with the 4,898 positive samples in the original samples identified by the first stage of annotation constitute the training set. The two-stage manual annotation costs 0.5\$ per sample. See Appendix~\ref{app:ann} for instructions for annotators.

\section{Experiments}
In our experiments, we evaluate the dataset using two approaches. First, we assess the zero-shot performance of state-of-the-art LLMs on the development and test sets. This aims to highlight the challenge posed by the test samples. Second, due to resource constraints, we conduct parameter-efficient fine-tuning on smaller models using the training data. This focuses on demonstrating the effectiveness of the training samples.

\subsection{Settings}

State-of-the-art LLMs that we use for zero-shot performance tests include GPT-4o \cite{openai2024gpt4technicalreport}, Qwen2.5-Plus \cite{qwen2024qwen25technicalreport}, and DeepSeek-v3 \cite{deepseekai2024deepseekv3technicalreport}. We provide the sample to the LLMs and ask for their judgment. The relatively small language models we use for training include Llama-3.1-8B \cite{grattafiori2024llama3herdmodels}, Mistral-7B \cite{jiang2023mistral7b}, and Qwen2.5-7B \cite{qwen2024qwen25technicalreport}. The parameter-efficient fine-tuning method we use is LoRA \cite{DBLP:conf/iclr/HuSWALWWC22}. See Appendix~\ref{app:detail} for training details. We use accuracy as the metric. Since there are equal numbers of positive and negative samples, the accuracy is equivalent to the commonly used balanced accuracy \cite{DBLP:journals/corr/abs-2303-15621}, which is the average of the accuracy on positive and negative samples. We also report the accuracy of positive and negative samples separately.

\begin{table}[]
\centering
\begin{tabular}{l|ccc}
\hline
\textbf{Dev}    & \textbf{Acc}  & \textbf{Acc}$_\text{p}$ & \textbf{Acc}$_\text{n}$ \\ \hline
GPT-4o          & \textbf{83.7} & 97.0            & 70.4            \\
Qwen2.5-Plus       & 81.6          & 97.0            & 66.2            \\
DeepSeek-v3     & 69.4          & 99.2            & 39.6            \\ \hline
Llama-3.1-8B    & \textbf{91.4} & 91.6            & 91.2            \\
Mistral-7B & 89.5          & 91.2            & 87.8            \\
Qwen2.5-7B      & 91.2          & 95.0            & 87.4            \\ \hline
\textbf{Test}   & \textbf{Acc}  & \textbf{Acc}$_\text{p}$ & \textbf{Acc}$_\text{n}$ \\ \hline
GPT-4o          & \textbf{83.9} & 96.2            & 71.6            \\
Qwen2.5-Plus       & 81.2          & 94.8            & 67.6            \\
DeepSeek-v3     & 69.4          & 99.4            & 39.4            \\ \hline
Llama-3.1-8B    & \textbf{90.6} & 90.4            & 90.8            \\
Mistral-7B & 89.8          & 92.0            & 87.6            \\
Qwen2.5-7B      & 88.5          & 90.4            & 86.6            \\ \hline
\end{tabular}
\caption{\label{tab:res} Results of experiments on the dev set and the test set. We report overall accuracy (\textbf{Acc}), accuracy on positive samples (\textbf{Acc}$_{\text{p}}$), and accuracy on negative samples (\textbf{Acc}$_{\text{n}}$) in percentage form.}
\end{table}

\subsection{Results}

Table~\ref{tab:res} reveals significant differences in performance between LLMs tested under zero-shot conditions and smaller models fine-tuned with parameter-efficient methods. Among the zero-shot LLMs, GPT-4o achieved the highest overall accuracy, outperforming Qwen2.5-Plus and DeepSeek-v3. However, even GPT-4o struggled with negative samples, achieving only 70.4\% accuracy on the dev set and 71.6\% on the test set. This limitation highlights a persistent challenge in distinguishing negative cases, which significantly impacts overall accuracy. DeepSeek-v3, while demonstrating near-perfect accuracy on positive samples, performed poorly on negative samples (39.6\% dev, 39.4\% test), indicating a clear trade-off between the two categories.

In contrast, smaller models fine-tuned with the training set achieved remarkable improvements, particularly in handling negative samples. Llama-3.1-8B stood out as the top performer, achieving 91.4\% accuracy on the dev set and 90.6\% on the test set, while maintaining a strong balance between positive and negative samples. These results suggest that the training data effectively addressed the challenges posed by negative samples, enabling the fine-tuned models to achieve significantly higher overall accuracy. Overall, the results underscore the effectiveness of fine-tuning in improving model robustness, particularly for negative samples. The dataset’s training data appears to play a crucial role in enhancing model performance, as evidenced by the fine-tuned models’ ability to achieve high accuracy across both positive and negative samples. These insights suggest that tailored training strategies and targeted fine-tuning can significantly enhance model capabilities, even for smaller models.

\section{Conclusion}
 
In this work, we propose the first large-scale Chinese dataset \textsc{CiteCheck} for citation faithfulness detection. To solve the high-cost problem caused by the lack of negative samples when constructing the dataset using strong RAG systems, we propose the method of data augmentation with two-stage manual annotation. This method allows us to construct a dataset with a balanced number of positive and negative samples at a relatively low cost and guarantees the quality of the dataset. We conduct experiments and validate the quality of the dataset in two aspects: (1) the test samples consisting of the original samples are challenging for detection, and (2) the training samples consisting of the original positive samples and the augmented negative samples can be effectively applied for training.

\section*{Limitations}
The main limitation of the dataset is the availability of only binary judgment labels (positive or negative). We do not manually label which part of the statement in the negative sample is unsupported, nor do we manually label the evidence in the documents that the statement in the positive sample is supported. However, key segments labeling and modifications in the LLM augmentation phase are available, which compensates for the limitation to some extent.

The main limitation of the experiments is the lack of more experiments on other test sets for the model obtained from training to show the generalization performance. This limitation comes from the lack of relevant Chinese datasets. We will continue to track the relevant Chinese datasets proposed and conduct experiments.

\section*{Ethics Statement}
\par We comply with the license to use language models for scientific research purposes only. Questions are collected with the permission of the license of open-source datasets or with the consent of the relevant users. The datasets we construct will also be open source for scientific research purposes. We conduct checks to minimize potential risk issues with datasets, including personal privacy concerns and harmful content.
\par The AI assistant we use in our work is Copilot (for simple code completion).

\bibliography{custom}

\clearpage
\appendix

\section{Unexpected Questions}
\label{app:question}
Real-world questions do not always have the correct premises. For example, in the question "\begin{CJK}{UTF8}{gbsn}水俣病的传染途径是什么？\end{CJK}(What is the route of infection for Minamata disease?)", Minamata disease is not an infectious disease. Taking this situation into account, we add a small number of human-written questions with incorrect premises and LLM-generated questions with hard-to-verify premises in the question collection phase. The number of these questions in the total number of questions is about 3\%.

\section{Prompt for LLM Augmentation}
\label{app:aug}
\begin{table*}

\centering

\begin{tabular}{|p{\textwidth}|}
\hline
\\ [2pt]
\begin{CJK}{UTF8}{gbsn}
\par 这里有一段陈述和对应的一段参考文本。请按如下步骤完成任务，严格按我给出的格式进行输出：
\par（1）找到参考文本中所有直接支撑陈述中信息的原始关键文段（可能有多个，每一处都要找到）。每行输出一个原始关键文段及其直接支撑的陈述中的信息，格式为“关键文段{编号}：{关键文段}（支撑陈述中的信息：{支撑信息}）”。
\par（2）请将关键文段分组，每组包含的关键文段支撑陈述中的相同或相关的信息，输出一行分组结果，格式为“关键文段分组：第一组：（第一组关键文段编号），第二组：（第二组关键文段编号）...”。例如，陈述中有2个信息，关键文段1支撑信息1，关键文段2支撑信息2，关键文段3支撑信息1，那么输出“关键文段分组：第一组：（1，3），第二组：（2）”
\par（3）选择一组关键文段，对其中支撑陈述中信息的部分进行修改，满足以下要求：
\par - 修改应该使得关键文段无法完全支撑陈述中的对应信息。
\par - 修改应该保持关键文段的逻辑通顺、关键文段中的信息之间不矛盾。
\par - 修改之后的关键文段应该在参考文本的上下文语境中保持逻辑通顺，且与参考文本中的其他内容不矛盾。
\par - 只修改支撑陈述中信息的部分，其它部分保持不变。
\par - 如果一组中有多个关键文段，修改后它们的信息应该保持一致。
\par 你需要尝试两种修改方法：
\par - 改变信息：将关键文段中的某一处信息修改为另外的信息。请不要进行与原信息产生直接冲突的修改。例如，原信息为“奥迪A7旗舰版的最高速度比上一代快”，合适的修改是“奥迪A7豪华版的最高速度比上一代快”，不合适的修改1是“奥迪A7旗舰版的最高速度比上一代慢”（使用反义词，与原信息直接冲突），不合适的修改2是“奥迪A7旗舰版的最高速度不比上一代快”（添加否定词，与原信息直接冲突）。
\par - 删除信息：将关键文段中的某一处信息删除。关键文段如果是完整的句子，删除信息后应该仍然是一个完整的句子。例如，原文段为“由于天气原因，项目推迟至3月15日启动”（完整的句子），合适的修改是“由于天气原因，项目推迟至3月启动”（仍然是完整的句子），不合适的修改是“由于天气原因”（不再是完整的句子）。
\par 对每种方法，输出被修改的关键文段，并检查其逻辑通顺程度，给出一个1\textasciitilde 10以内的整数作为评分（越高表示越通顺）。每行输出一个修改后的关键文段，格式为“{方法}-修改后的关键文段{编号}：{修改后的关键文段}（逻辑通顺程度：{分数}）”。
\end{CJK} 
\\ [5pt]
\\ [5pt]
\hline

\end{tabular}

\caption{\label{tab:prompt} The complete prompt for the LLM augmentation.}
\end{table*}

\begin{table*}

\centering

\begin{tabular}{|p{\textwidth}|}
\hline
\\ [2pt]
\par Here is a statement and a corresponding piece of reference text. Please complete the task as follows, strictly following the format I have given for the output:
\par (1) Find all the original key passages in the reference text that directly support the information in the statement (there may be more than one, find each one). Output one original key passage per line and the information in the statement it directly supports in the format “Key passage {number}: {key passage} (information in the supporting statement: {supporting information})”.
\par (2) Please group key passages, each group contains key passages supporting the same or related information in the statement, output one line of the grouping results in the format of “Key passage grouping: Group 1: (first group of key passage numbers), Group 2: (second group of key passage numbers) ...”. For example, if there are 2 pieces of information in the statement, key paragraph 1 supports information 1, key paragraph 2 supports information 2, and key paragraph 3 supports information 1, then the output is “Key Paragraph Grouping: Group 1: (1, 3), Group 2: (2)”.
\par (3) Select a group of key text segments and modify the parts of them that support the information in the statement to meet the following requirements:
\par - The modification should make it impossible for the key passage to fully support the corresponding information in the statement.
\par - The modifications should maintain the logical flow of the key passages and no contradictions between the information in the key passages.
\par - The modification should keep the key paragraph logically coherent in the context of the reference text and not contradict the rest of the reference text.
\par - Modify only the parts that support the information in a statement, leaving the rest unchanged.
\par - If there is more than one key passage in a set, the information in them should remain consistent after revision.
\par You need to try two methods of modification:
\par - Changing the message: modifying the message in one part of the key paragraph to another. Do not make changes that directly conflict with the original information. For example, if the original message is “The Audi A7 Signature Edition has a faster top speed than its predecessor”, an appropriate change would be “The Audi A7 Luxury Edition has a faster top speed than its predecessor”, and an inappropriate change1 would be “The Audi A7 Signature Edition has a slower top speed than its predecessor” (using an antonym, which is in direct conflict with the original message), and inappropriate modification 2 is ‘The top speed of the Audi A7 Signature Edition is not faster than the previous generation’ (adding a negative word, which is in direct conflict with the original message).
\par - Delete Information: Remove information from a place in a key paragraph. If the key paragraph is a complete sentence, it should still be a complete sentence after deleting the information. For example, if the original paragraph reads “Due to weather conditions, the project was delayed until March 15” (complete sentence), an appropriate change would be “Due to weather conditions, the project was delayed until March” (still a complete sentence), an inappropriate change would be “Due to the weather” (no longer a complete sentence).
\par For each method, output the key passage that was modified and check its logical fluency, giving an integer within 1 to 10 as a rating (higher means more fluent). Output one modified key passage per line in the format “{method}-modified key passage {number}: {modified key passage} (logical fluency: {score})”.
\\ [5pt]
\\ [5pt]
\hline

\end{tabular}

\caption{\label{tab:prompt_en} The complete prompt for the LLM augmentation (translated into English).}
\end{table*}

See Table~\ref{tab:prompt} for the prompt for LLM augmentation. Table~\ref{tab:prompt_en} provides an English version.

\section{Instructions for Annotators}
\label{app:ann}
\subsection{First Stage}
\begin{figure*}
    \centering
    \includegraphics[width=0.3\textwidth]{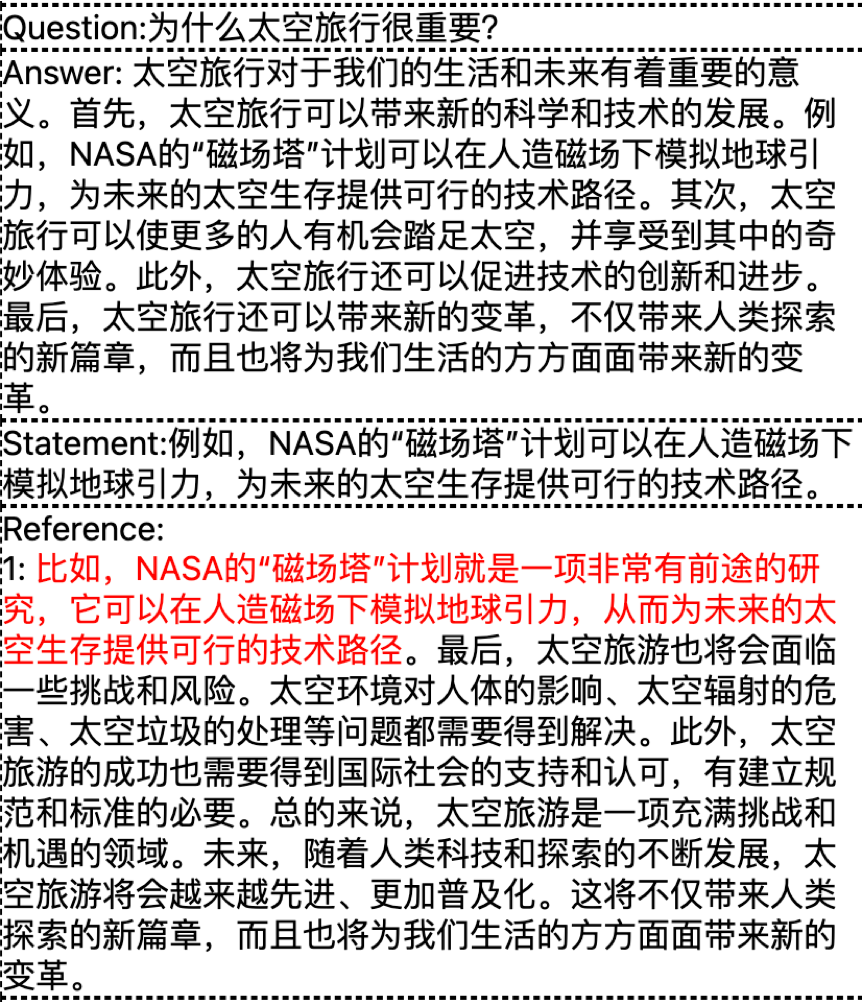}
    \includegraphics[width=0.3\textwidth]{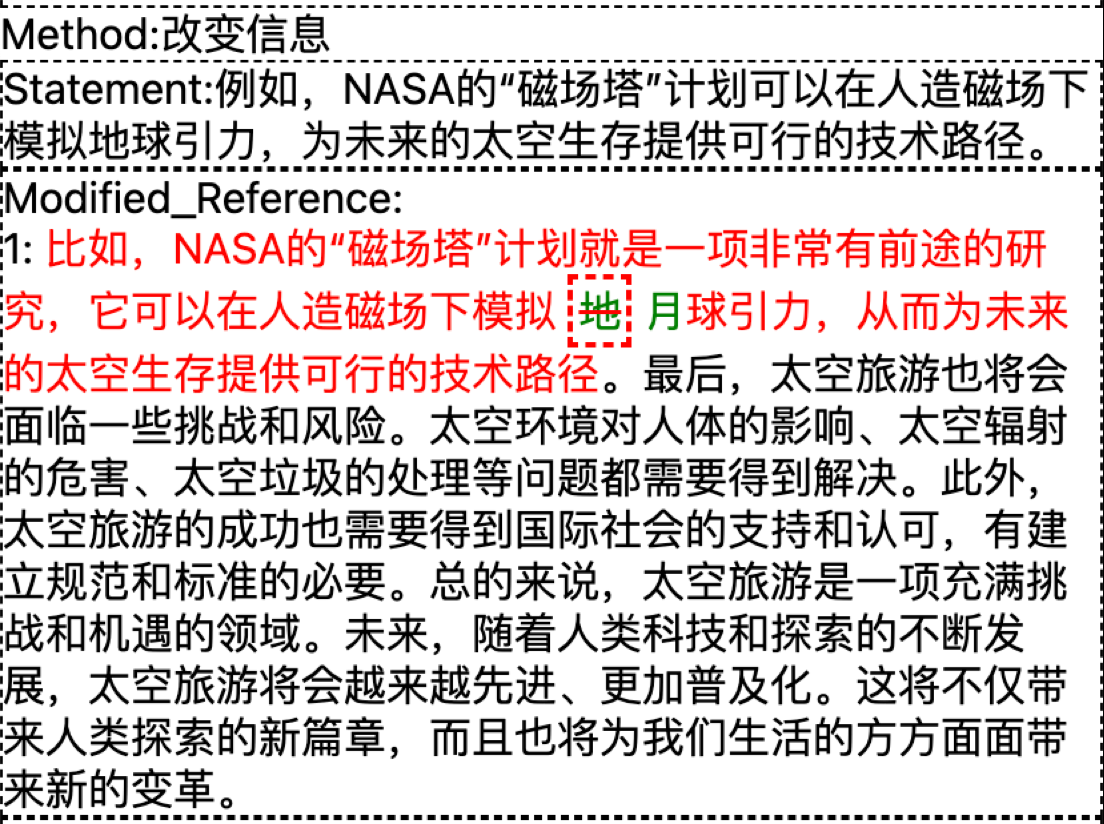}
    \includegraphics[width=0.3\textwidth]{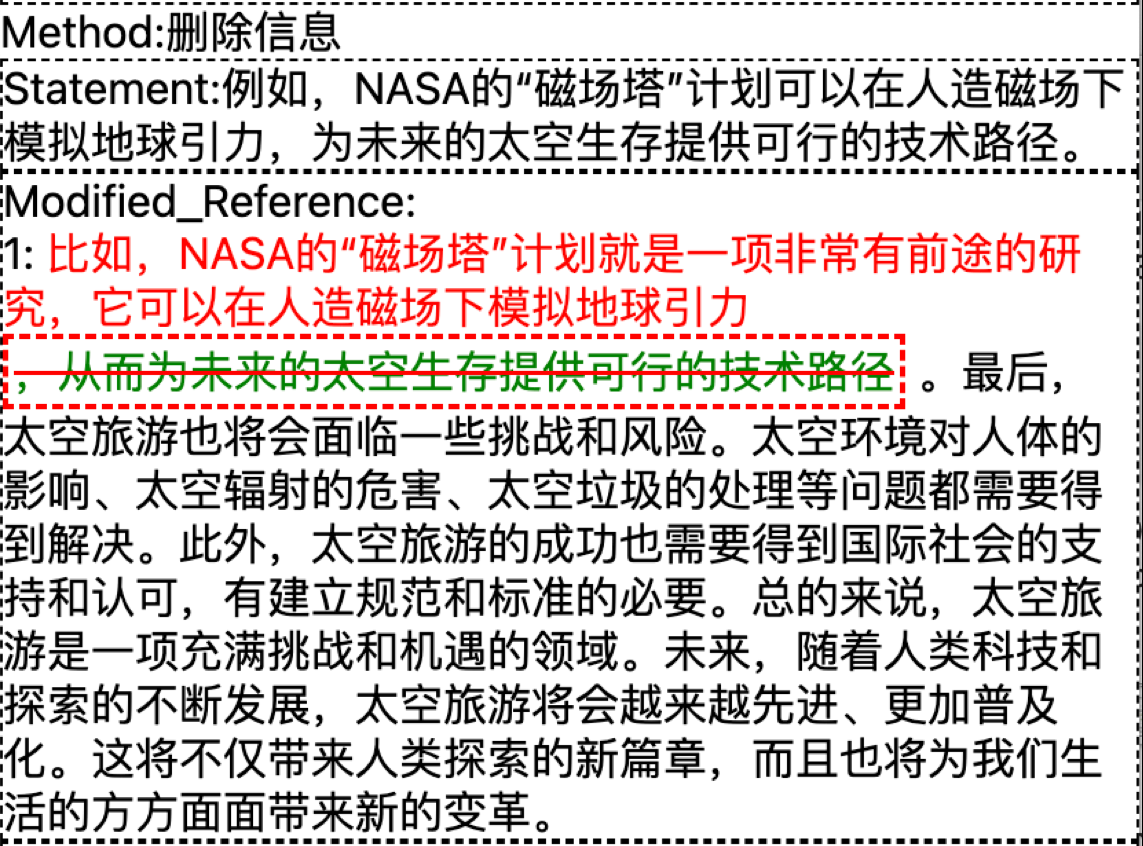}
    \caption{Examples of interfaces that provide samples to the annotators. The first figure shows an example of the first stage. The last two images show the second stage with the same sample modified (information changed/deleted).}
    \label{fig:stage}
\end{figure*}

In the first stage, we provide the annotators with the question, answer, statement, and cited documents. What LLM considers to be key segments are highlighted in red in the cited documents (see Figure~\ref{fig:stage} for an example). We instruct the annotators to follow the process below:

\par (1) First look at the highlighted text. If the highlighted text fully supports the statement, then the annotation is positive; if the highlighted text contradicts the statement, then the annotation is negative.

\par (2) If the annotation cannot be derived from the highlighted text, then look at the rest of the documents to make the annotation. When the documents fully support the statement, the label is positive, and when there is any information in the statement that contradicts the documents or information that is not mentioned in the documents, the label is negative.

\subsection{Second Stage}

In the second stage, we provide the annotator with the statement and the modified documents. In the documents, the modified parts are highlighted in green, where the dashed and crossed-out text is deleted and the rest is added (see Figure~\ref{fig:stage} for examples). 

For the annotation of whether the quality of the modification is acceptable, the annotators are instructed to note that qualified modifications need to satisfy the following two requirements: (1) There are no contradictions within each modified document. (2) The modified key segments are fluent in their own right and in the context of the document. The annotation for support is the same as the first stage, but based on the modified documents.

\section{Input and Training Details}
\label{app:detail}
We input the statement and the cited documents into the model and ask the model to determine whether the statement is fully supported by the documents, outputting yes or no. For input, we label and concatenate the cited documents in order (as shown in Table~\ref{tab:dataset}). For training, we use the following settings: For training, we use the following settings: learning rate is 5e-4, number of epochs is 10, scheduler is cosine scheduler, warmup ratio is 0.03, batch size is 256, and LoRA setting is $r=8$, $a=32$ and 0.1 dropout. We report the model performance for the epoch that achieves the best performance on the dev set.
\label{app:detail}

\section{Related Works}
Language models are known to produce hallucinations - statements that are inaccurate or unfounded~\citep{MaynezNBM20,HuCLGWYG24}. To address this limitation, recent research has focused on augmenting LLMs with external tools such as retrievers~\citep{GuuLTPC20,BorgeaudMHCRM0L22,LiuCtrla2024} and search engines~\citep{WebGPT2021, Komeili0W22, TanGSXLFLWSLS24}. While this approach suggests that generated content is supported by external references, the reliability of such attribution requires careful examination. Recent studies have investigated the validity of these attributions. \citet{DBLP:conf/emnlp/LiuZL23} conducted human evaluations to assess the verifiability of responses from generative search engines. \citet{hu2024evaluate} further investigate the reliability of such attributions when giving adversarial questions to RAG systems. Their findings revealed frequent occurrences of unsupported statements and inaccurate citations, highlighting the need for rigorous attribution verification~\citep{RashkinNLA00PTT23}. However, human evaluation processes are resource-intensive and time-consuming. To overcome these limitations, existing efforts~\citep{GaoDPCCFZLLJG23,DBLP:conf/emnlp/GaoYYC23} proposed an automated approach using Natural Language Inference models to evaluate attribution accuracy. While several English-language benchmarks have been developed for this purpose~\citep{DBLP:conf/emnlp/YueWCZS023}, comparable resources in Chinese are notably lacking. Creating such datasets presents unique challenges, particularly in generating realistic negative samples (unsupported citations).  To address this gap, we introduce the first large-scale Chinese dataset for citation faithfulness detection, developed through a cost-effective two-stage manual annotation process.

\end{document}